%% file: main.tex
\documentclass{article}
\usepackage{arxiv}
\input epsf
\usepackage{amsmath,amssymb,amsfonts}
\usepackage{algorithmic}
\usepackage{graphicx}
\usepackage{textcomp}
\usepackage{xcolor}
\usepackage{comment}
\usepackage{subcaption}
\usepackage{multirow}
\usepackage{multicol}
\usepackage{fixltx2e}
\usepackage{relsize}
\usepackage{mathbbol}
\usepackage{url}
\usepackage{hyperref}

\newcommand{\JB}[1]{\textcolor{red}{JB: #1}}

\begin{document}

\title{\LARGE Analyzing the impact of climate change on critical infrastructure from the scientific literature: A weakly supervised NLP approach}  

\author{
  Tanwi Mallick \\
  Mathematics and Computer Science Division\\
  Argonne National Laboratory, Lemont, IL \\
  \texttt{tmallick@anl.gov} \\
   \And
   Joshua David Bergerson \\
  Decision and Infrastructure Sciences Division \\
  Argonne National Laboratory, Lemont, IL \\
  \texttt{jbergerson@anl.gov} \\
  \And
  Duane R. Verner \\
  Decision and Infrastructure Sciences Division  \\
  Argonne National Laboratory, Lemont, IL \\
  \texttt{dverner@anl.gov} \\
  \And
  John K Hutchison\\
  Decision and Infrastructure Sciences Division  \\
  Argonne National Laboratory, Lemont, IL \\
  \texttt{jhutchison@anl.gov} \\
  \And
  Leslie-Anne Levy\\
  Decision and Infrastructure Sciences Division  \\
  Argonne National Laboratory, Lemont, IL \\
  \texttt{llevy@anl.gov} \\
  \And
  Prasanna Balaprakash \\
  Mathematics and Computer Science Division \& 
  Argonne Leadership Computing Facility \\
  Argonne National Laboratory, Lemont, IL \\
  \texttt{pbalapra@anl.gov} 
}





\maketitle
\thispagestyle{plain}
\pagestyle{plain}

\begin{abstract}

Natural language processing (NLP) is a promising approach for analyzing large volumes of climate-change and infrastructure-related scientific literature. However, best-in-practice NLP techniques require large collections of relevant documents (corpus). Furthermore, NLP techniques using machine learning and deep learning techniques require labels grouping the articles based on user-defined criteria for a significant subset of a corpus in order to train the supervised model. Even labeling a few hundred documents with human subject-matter experts is a time-consuming process. To expedite this process, we developed a weak supervision-based NLP approach that leverages semantic similarity between categories and documents to (i) establish a topic-specific corpus by subsetting a large-scale open-access corpus and (ii) generate category labels for the topic-specific corpus. The developed approach performs programmatic labeling with weak supervision by defining labeling functions to capture semantic similarity, rather than syntactic or rule-based similarity, between documents and categories. 
To support this approach, we developed climate hazard definitions based on several authoritative sources on climate change and used definitions developed by the U.S. Department of Homeland Security for National Critical Functions (NCFs) that encompass the functions provided by the critical infrastructure. By comparing the semantic similarity between documents in the Semantic Scholar Open Research Corpus (S2ORC), we established a large corpus of 17,136 research articles focused on both climate change hazards and NCFs as a subset of S2ORC. We use multiple text-embedding models as weak learners within the programmatic labeling technique. To speed up the labeling process, we scaled the weak supervision labeling across multiple GPUs on a computing cluster at the Argonne Leadership Computing Facility. 
In comparison with a months-long process of subject-matter expert labeling, we assign category labels to the whole corpus for 55 categories of critical infrastructure functions and 18 categories of climate-related hazards using weak supervision and supervised learning in about 13 hours. The labeled climate and NCF corpus enable targeted, efficient identification of documents discussing a topic (or combination of topics) of interest and identification of various effects of climate change on critical infrastructure, improving the usability of scientific literature and ultimately supporting enhanced policy and decision making. To demonstrate this capability, we conduct topic modeling on pairs of climate hazards and NCFs to discover trending topics at the intersection of these categories. This method is useful for analysts and decision-makers to quickly grasp the relevant topics and most important documents linked to the topic.



\end{abstract}

\keywords{Weak supervision, BERT embedding, topic modeling, climate hazard, critical infrastructure}

\section{Introduction}
\input{intro}

\section{Related work}

\input{related}

\section{Overview of climate change hazard and NCF categories}
\input{climate}

\section{Methodology}
\input{methodology}

\section{Results}
\input{results}

\section{Conclusion}
\input{conclusion}

\section*{Acknowledgments}
This material is based in part upon work supported by the Laboratory Directed Research and Development (LDRD), Argonne National Laboratory. 
This research used resources from the Argonne Leadership Computing Facility, which is a DOE Office of Science User Facility under contract DE-AC02-06CH11357. 

\bibliographystyle{IEEEtran}
\bibliography{reference.bib}

\section{Appendix}
\input{appendix}

\section*{Government license}
The submitted manuscript has been created by UChicago Argonne, LLC, Operator of Argonne National Laboratory ("Argonne"). Argonne, a U.S. Department of Energy Office of Science laboratory, is operated under Contract No. DE-AC02-06CH11357. The U.S. Government retains for itself, and others acting on its behalf, a paid-up nonexclusive, irrevocable worldwide license in said article to reproduce, prepare derivative works, distribute copies to the public, and perform publicly and display publicly, by or on behalf of the Government.  The Department of Energy will provide public access to these results of federally sponsored research in accordance with the DOE Public Access Plan. http://energy.gov/downloads/doe-public-access-plan.

\end{document}

%% file: intro.tex
Critical infrastructure systems throughout the United States and the world are increasingly at risk due to intensifying natural hazards driven by anthropogenic climate change. 
The effects of climate change, such as more frequent extreme weather events, sea level rise, and temperature fluctuations, can degrade or disrupt critical infrastructures such as roads, bridges, airports, ports, and power plants and disrupt the supply of essential goods and services.
These disruptions may threaten the safety and well-being of communities, cause significant financial losses, and  inflict long-term socioeconomic damage. Research on climate change, its effects on critical infrastructure, and adaptation strategies is constantly evolving and published at a blistering pace, reducing the ability of decision-makers to review, synthesize findings, and apply them in their climate adaptation decision-making processes. Decision-makers require actionable, understandable guidance on  the potential impacts of different climate conditions on critical infrastructure systems and
 climate change adaptation strategies to enhance the resilience of these systems and reduce their future disruptions. 
Given the overwhelming volume of publications on these topics, decision-makers  and policy makers face significant challenges in identifying optimal strategies for adapting critical infrastructure. 

Artificial intelligence (AI) techniques, including natural language processing (NLP), machine learning, and topic modeling, can efficiently canvas the existing literature to improve climate adaptation policy- and decision-making. Unsupervised learning \cite{callaghan2020topography, varini2020climatext, rolnick2022tackling} is commonly used by researchers to find groups or clusters in a large corpus of documents. The process can be challenging, however, in the case of a large corpus of noisy and unstructured data. Moreover, unsupervised learning is unable to predict specific classes or categories because it does not employ labeled training examples. 
In practice, infrastructure professionals such as civil engineers, construction managers, and urban planners are interested in reviewing documents associated with a specific topic of interest (i.e., user-defined class) rather than arbitrary clusters or groups. 
For example, to design and build infrastructure with greater resilience against extreme weather events, a civil engineer could want to explore how specific climate events such as droughts and extreme rainfall may impact wastewater systems.
Hence, it is crucial to identify documents belonging to specific  categories, such as droughts, extreme rainfall, supply water, and emergency management. Supervised learning enables this categorization of documents through user-defined classes.
One of the major challenges impeding use of supervised learning is the necessity of a labeled dataset for training the model. When a labeled dataset is not readily available, manual labeling with the help of domain experts is generally required, a process that can be time-consuming and laborious. Moreover, the involvement of multiple labelers can lead to discrepancies and inconsistencies 
in labels. Alternatively, the amount of labeling required can be decreased by using an active-learning method \cite{gao2020consistency, goudjil2018novel, callaghan2021machine} by finding the texts that best inform a supervised learning algorithm. Active learning chooses unlabeled documents for which the learned supervised model has the most uncertainty (or, thought of another way, the least confidence in predicted labels). However, this approach can become a bottleneck because of the need for human experts in the loop. 

To that end, we developed a powerful and effective method to categorize a large number of documents in a limited time and identify the topics for a specific category. Our goal is to  help community leaders around the world navigate the vast number of research papers on climate change and infrastructure impacts. 
We designed a novel method for programmatic labeling of a large corpus using weak supervision, allowing the dataset to be labeled with minimal human involvement. 
Programmatic labeling with weak supervision is accomplished by defining labeling functions that  capture semantic similarity instead of patterns, heuristics, or rules for labeling. To enable programmatic labeling of climate hazards with weak supervision, we define 18 climate hazard categories and extract category definitions based on a review of recent publications from several authoritative sources including the Intergovernmental Panel on Climate Change (IPCC), National Academies of Sciences, Engineering, and Medicine (NASEM), American Society of Civil Engineers (ASCE), the U.S. Global Change Research Program (USGCRP), and the National Oceanic and Atmospheric Administration (NOAA). To support programmatic labeling of critical infrastructure, we use existing definitions published by the Cybersecurity and Infrastructure Security Agency
as the set of 55 National Critical Functions (NCFs), which collectively represent the critical functions provided by critical infrastructure systems \cite{ncfdefinition}. 
The labeling functions in our weak supervision method evaluate the semantic similarity between the definitions and unlabeled documents. To increase performance by making the method more robust, we construct the labeling functions for the first time utilizing multiple semantic embedding techniques. The semantic embedding models have different architectures and  are pretrained on different general-purpose datasets. Therefore, a single embedding technique is not sufficient to label our climate and NCF dataset. Hence, we define multiple labeling functions using different embedding techniques to cover different aspects of the data. The weak supervision model generates the probabilistic labels by maximizing the overlap and minimizing the conflict among the multiple labeling functions using probabilistic modeling. 
Once labeled with weak supervision, the documents with the probabilistic labels 
serve as the training set for the supervised learning model. After the model is trained, it classifies the remainder of the corpus not included in the training set. We scaled the weak supervision and supervised learning methods across several GPUs to shorten the time required for labeling and categorizing the corpus.
The categorization of the corpus across sets of climate hazards and critical infrastructure functions 
enables users to quickly identify a subset of the corpus associated with a particular category of interest. 

The proposed method analyzes research published on climate change and its impacts on communities and infrastructure, distills it into specific categories, and finds the diverse topics discussed in the categories. 
We manually evaluate the documents belonging to the individual topics and qualitatively assess the performance of our proposed approach. The main contributions of the paper are as follows: 
\begin{itemize}
    \item We developed multilabel programmatic labeling, where a document could have a positive label for more than one category, using weak supervision with multiple  semantic embedding techniques as programmatic labeling functions.
    \item Using weak and supervised learning, we developed two domain-specific corpora (climate change hazards and critical infrastructure) by subsetting an open-source large-scale corpus.
    \item We scaled the weak and supervised learning over multiple GPUs and reduced the overall labeling time.
    \item For the first time, we investigated the various effects of particular climate hazards on particular NCFs using the topic model. We found a variety of 
    topics for climate hazards and NCF pairs because of categorization, which  allowed us to filter documents by categories.
    \item We demonstrated that our proposed method can effectively categorize the documents based on targeted 
    interests and elicit specific insights from the large corpus, which may inform decision-making and help communities prepare for climate change. 
\end{itemize}




%% file: related.tex
Various machine learning models have been used to analyze, model, and track the components of climate change. Regardless of methodological advances in the NLP domain, the use of NLP techniques to explore the impacts of climate change on critical infrastructure has received little attention in the literature. The research  currently available in this field can be divided into two classes:  unsupervised learning and supervised learning.



\subsection*{Unsupervised learning}
In the climate domain,  comparatively many attempts have focused on unsupervised learning or topic modeling. Hsu et al. \cite{hsu2021diverse} performed topic modeling on climate strategy documents across different countries, regions, cities, and companies. They discovered connections between regional and national strategies and identified gaps in climate actions. Benites et al. \cite{benites2018topic} carried out latent unsupervised topic modeling using Latent Dirichlet Allocation (LDA) to 
identify trending topics related to Brazilian ethanol production and its connection to food security and climate change. Chang et al. \cite{chang2021applying} identified distinct topic categories such as climate change, environmental education, and sustainability/sustainable development based on the topics and subtopics identified using LDA and K-means clustering and attempted 
to tag the papers in an unsupervised manner. The results show that unsupervised learning models produce diverse clusters of documents based on similarity or dissimilarity measures.  As is well documented in the literature, unsupervised learning cannot group documents into specific (i.e., predefined) categories such as climate hazards or critical infrastructure functions. 
Therefore, the application of unsupervised learning may be less informative than supervised learning for identifying information associated with a specific topic (or combination of topics) of interest, such as the risk wildfire poses to the Supply Water NCF. Conversely, models built with supervised learning  enable specific user groups to search for information associated with specific categories of interest.


\subsection*{Supervised learning}
Because of the scarcity of labeled data, use of supervised learning in the domains of climate change and critical infrastructure is very limited. A few papers describe previous efforts to label small datasets to support exploratory analysis. Marsi et al. \cite{marsi2014towards} attempted to automatically derive relationships between quantitative climate variables from unstructured text. In this work, texts were manually annotated in order to draw connections between various variables. Because of the high cost of human annotation, the researchers were unable to demonstrate their method on a large corpus. Chen et al. \cite{chen2019detecting} created a deep neural network classifier to recognize users who reject climate change based on the content of their tweets. This method, which was tested on a small corpus of 2,000 tweets, also involved manual tagging of the climatic material. Once the content was labeled, their model classified it with an accuracy of 88\%.


Overall, the review of available research suggests that no work has been done to analyze the impact of climate change on critical infrastructure from existing literature using supervised learning. Because of the lack of a publicly available labeled dataset, supervised learning has been used less frequently than unsupervised learning on climate corpora.

Our proposed approach of labeling a large corpus using weak supervision with multiple semantic embedding techniques is unique and, to the best of our knowledge, no one has ever used weak supervision to label climate and infrastructure corpora. Furthermore, we show that topic modeling on a corpus of documents with accompanying climate and infrastructure category labels enables insights that were previously unexplored because of the lack of labeled datasets.


%% file: climate.tex
To analyze how climate change hazards affect national critical functions, we took into account both of these domains in the current work. An overview of these two domains is given in the ensuing subsections.

\subsection{Climate Change Hazards}
\begin{figure}[!ht]
\centering
   \includegraphics[width=12cm]{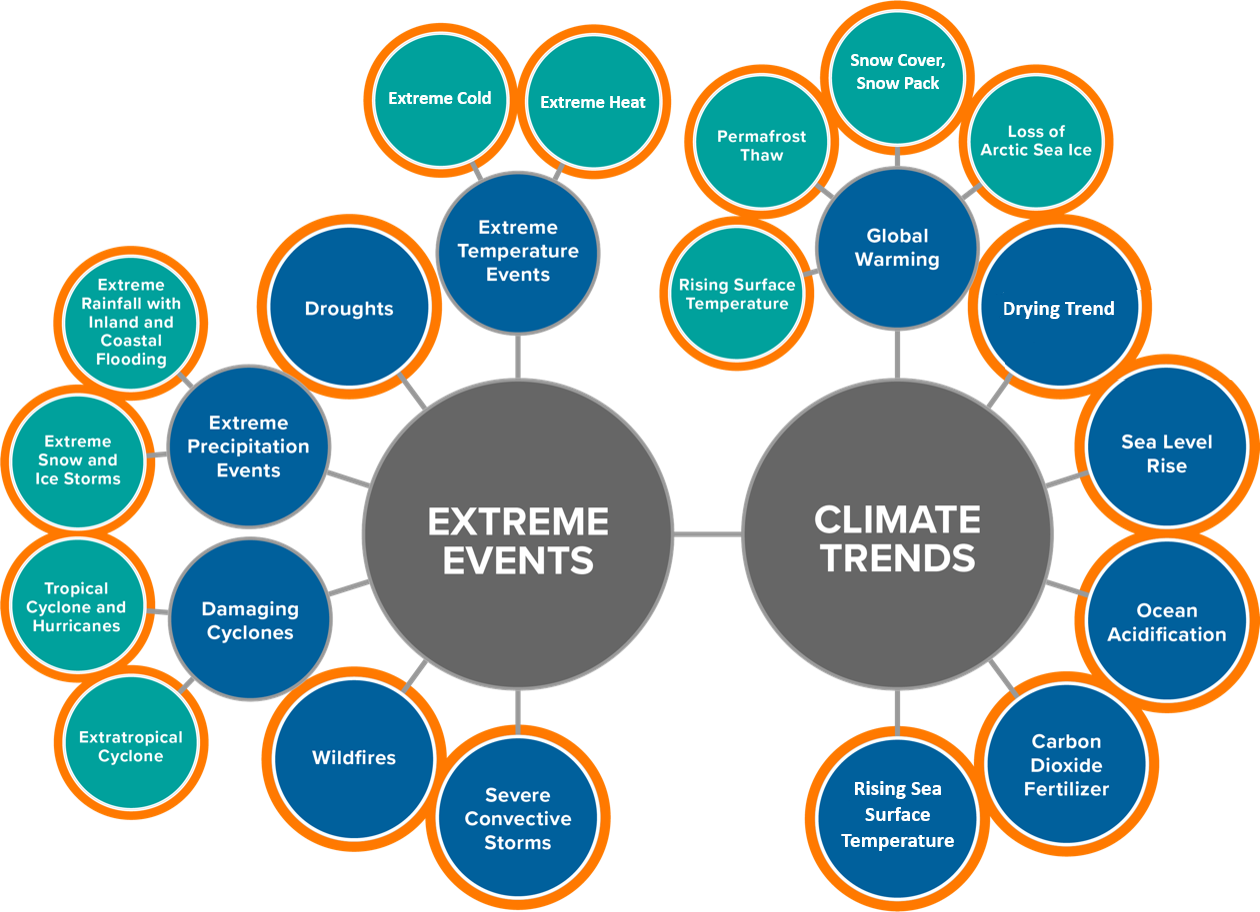}  
\caption{Hierarchical structure of climate change hazards, with orange rings identifying hazards considered in the NLP modeling } 
\label{fig_climate}
\end{figure}

To establish a set of climate change hazards (i.e., natural hazards projected to be altered by climate-change-driven effects and associated climate trends), we reviewed several authoritative data sources and reports on climate change hazards and impacts to critical infrastructure systems \cite{ipcc2014, national2016attribution, task2021impacts, usgcrp}. 
The USGCRP Fourth National Climate Assessment \cite{reidmiller2019fourth} and IPCC Sixth Assessment Report \cite{ipcc2022} document the current state of climate science, give future climate projections associated with potential future emissions scenarios using global climate models, potential socioeconomic impacts of future climate conditions on the environment, built infrastructure, and communities, and provide recommendations for policy and decision-makers.  NASEM explored the evolving science of attribution, examining the extent to which science supports quantifying the contribution of climate change to the occurrence (i.e., frequency, severity, duration) of extreme weather events \cite{national2016attribution}. As part of this effort, this work established categories of extreme weather events and characterized these categories based on the state of the science's understanding of the impact of climate change on these events and confidence in the attribution of these effects to climate change. The ASCE Task Committee on Future Weather and Climate Extremes of the Committee on Adaptation to a Changing Climate conducted an extensive literature review to explore uncertainty and identify consensus in the literature on projected future extreme weather and climate events considering the effects of anthropogenic climate change \cite{task2021impacts}. The committee also explored the likely impacts of future extreme weather and climate events on critical infrastructure systems and resilience considerations for climate adaptation.   
Based on the review of these sources, we identified a hierarchical structure of climate change hazards as shown in Figure \ref{fig_climate}. 
At the highest level, the hierarchy delineates extreme climate events and climate trends, where the former has a definitive start and end point (i.e., has a countable frequency and duration) while the latter captures shifts in average or seasonal climate conditions. Ultimately, 18 climate change hazards (identified with orange rings in Figure \ref{fig_climate}) were considered. We note that the 18 climate categories are not independent and that high pairwise correlation may exist between specific pairings of climate hazards (e.g., wildfire and drought). 

To enable the weak supervision approach (described later in the  manuscript), we identified and extracted definitions from the literature for each climate hazard. For this effort, we reviewed existing climate hazard definitions from several sources, including NOAA, USGCRP, and IPCC. Based on this review, we established definitions for each climate hazard.


\subsection{National Critical Functions}
A large volume of literature focuses on risks associated with climate change impacts on physical infrastructure systems and associated adaptation strategies for managing this risk. Given that critical infrastructure systems support nearly every aspect of modern society, this focus makes sense. However, this focus on built infrastructure stops short of identifying how disruption of critical infrastructure impacts the communities in which they are located. Further, the focus on physical infrastructure overlooks several key cyber and socioeconomic aspects of infrastructure operations. To broaden understanding of critical infrastructure and the ways in which it supports communities, the National Risk Management Center  in the  
Cybersecurity and Infrastructure Security Agency (CISA) led an effort to establish a functional decomposition of the 16 critical infrastructure sectors. This decomposition focused on identifying the functions that are deemed ``so vital to the United States that their disruption, corruption, or dysfunction would have a debilitating effect on security, national economic security, national public health or safety, or any combination thereof" \cite{ncf}. Resulting from this work, CISA published a set of 55 NCFs in April 2019 that encapsulate the critical functions of the government and private sector.  The NCFs capture how critical infrastructure supports communities, enabling improved risk management of critical infrastructure across communities throughout the nation. The 55 NCFs are classified into four categories: connect, distribute, manage, and supply. The 55 NCFs under the four broad categories are listed in the Appendix \ref{sec_ncfs}.

%% file: methodology.tex
\begin{figure*}[!ht]
\centering
   \includegraphics[width=\linewidth]{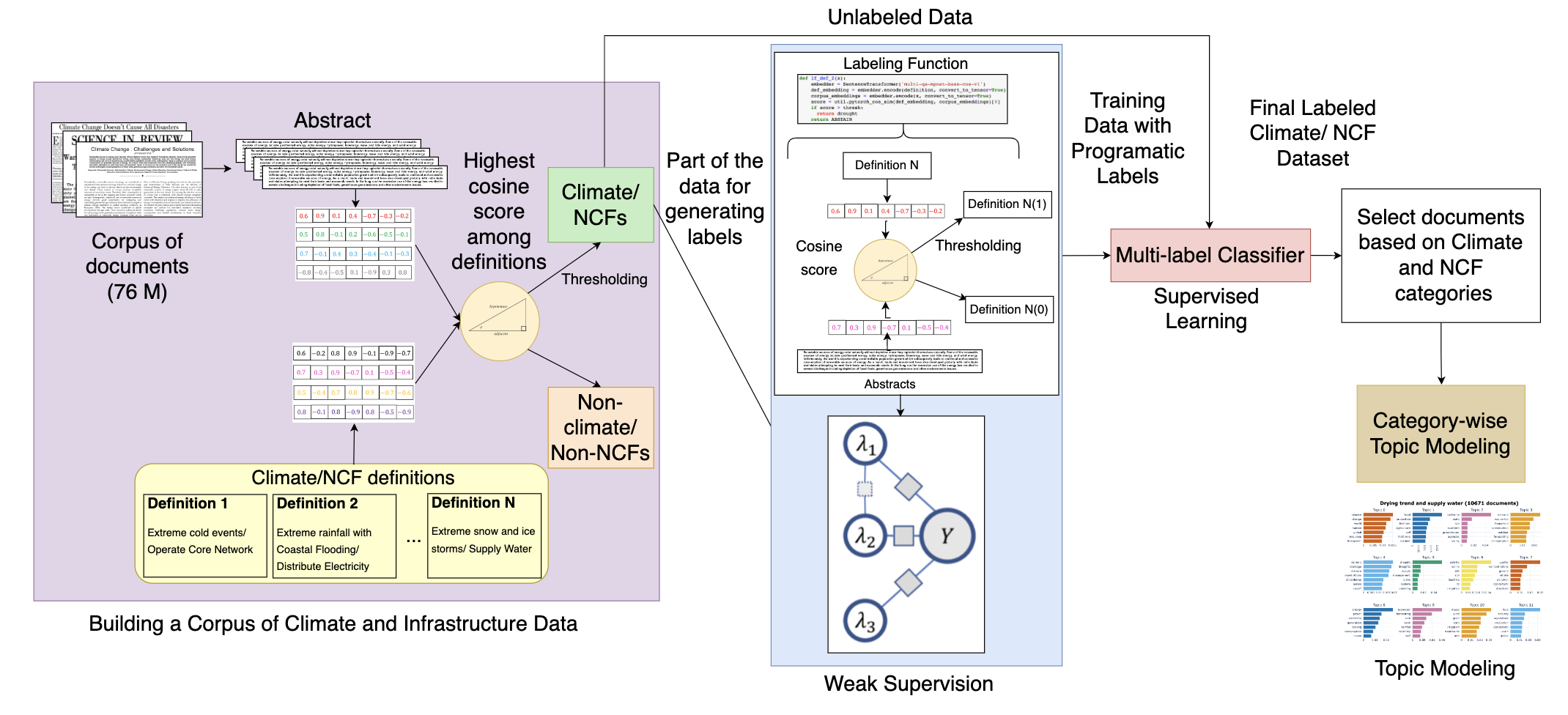}  
\caption{Visual representation of our approach in step-by-step manner}
\label{fig_approach}
\end{figure*}

In this paper we study the impact of climate change on critical infrastructure to support informed decision-making. To analyze the impact and elicit specific insights from the large corpus, we propose the methodology depicted graphically in Figure \ref{fig_approach}. Our proposed methodology consists of four steps: (1) building of climate and infrastructure corpus, (2) weak supervision, (3) supervised learning, and (4) category-wise topic modeling. The subsequent sections include detailed explanations of each step.

 

\subsection{Building the Climate and Infrastructure Corpus}\label{sec_preproc}



We built our climate and NCF corpus from S2ORC \cite{lo-wang-2020-s2orc}, a general-purpose corpus for NLP and text mining research. By combining data from several different sources and identifying open-access publications, this corpus comprises 136 million paper nodes with 76 million abstracts and 12.7 million full-text papers. The corpus covers diverse fields  including medicine, biology, chemistry, engineering, computer science, physics, materials science, math, psychology, economics, political science, business, geology, sociology, geography, environmental science, art, history, and philosophy.

As the first step to building our climate and NCF corpus, we extract climate-related documents from S2ORC. The step-by-step process of document extractions is described in Figure \ref{fig_approach}. Given the abstracts from S2ORC and definitions for our 18 climate hazard categories, we compute semantic embeddings using a pretrained sentence transformer \cite{reimers-2020-multilingual-sentence-bert}, a Python framework for sentence and paragraph embedding that employs bidirectional encoder representations from transformers (BERT) \cite{devlin2018bert}. 
Next, we compare the semantic similarity between the abstract embeddings and the embeddings of climate definitions to determine which documents discuss climate-related issues.
We use cosine similarity (CS) \cite{reimers2019sentence} to compare each abstract embedding with each definition embedding.  Given two embedding vectors $x$ and $y$, CS can be calculated as
\begin{equation*}
    CS = \frac{x.y}{||x|| ||y||},
\end{equation*}
where the dot product $x.y = \sum_{i=1}^n x_i y_i$, $n$ is the length of vectors, the norm of the vector $x$ is $||x|| = \sqrt{\sum_{i=1}^n x_i^2}$, and the norm of the vector $y$ is $||y|| = \sqrt{\sum_{i=1}^n y_i^2}$.  The cosine similarity score ranges from -1 to 1, with higher scores indicating stronger similarity between the two vectors. For each abstract $j$, we determine the maximum CS score between the abstract embedding and the embedding of each climate hazard category as $CS_{j_{max}} = max(CS_{j1}, CS_{j2}, ..., CS_{jN})$, where $N$ is the number of categories.  We define a rule-based classifier based on the maximum CS score for an abstract to determine whether or not an abstract is climate-related, as follows:

\begin{equation}
  ClimateFlag_{j} =\begin{cases}
    1, & \text{if $CS_{j_{max}}>threshold$}.\\
    0, & \text{otherwise}.
  \end{cases}
\end{equation}
If the $CS_{j_max}$ score for an abstract is above a certain threshold, the document is considered climate-related and  is assigned a climate flag of 1; if the score is below the threshold, the document is considered non-climate and is assigned a climate flag of 0. This process establishes a climate-focused corpus as a subset of S2ORC.

Following the extraction of documents related to climate change from S2ORC, we similarly extract documents related to NCFs from the climate-focused corpus. We compute semantic embeddings using the same pretrained sentence transformer based on the abstracts and NCF definitions. The semantic similarity between the abstracts and the NCF definitions is then calculated by using cosine similarity to determine whether or not the documents are discussing NCF-related topics. This two-step process allows us to define a corpus of documents focused on our climate change hazard categories and NCF categories as a subset of S2ORC.






\subsection{Weak Supervision} \label{sec_ws}

Here we discuss the weak supervision \cite{ruhling2021end} approach we adopt to programmatically label the large corpus of documents focused on climate change hazards and NCFs with minimum manual supervision. Weak supervision is a branch of machine learning where multiple domain heuristics or knowledge-based imprecise rules, which are sometimes referred to as labeling functions (LFs) \cite{bach2017weaksupervision}, are used to come up with a true label of the dataset. This method eliminates the cost and difficulty of getting hand-labeled datasets, which can be prohibitively expensive or sometimes unfeasible. Furthermore, manual labeling is  rigid; changes in the categories often require relabeling of a dataset. Weak supervision is far more flexible, as changes to the categories require only minimal effort to update LFs. The LFs determine whether a document should belong to a particular category or not. The weak supervision technique generates probabilistic labels by maximizing the coverage and minimizing the conflict across LFs. In this paper, we use the weak supervision package Snorkel \cite{ratner2017snorkel} to label our dataset.

\subsubsection{Weak supervision using Snorkel} 
The training of machine learning models using weak supervision involves three main steps.

First, defining the labeling function is one of the key components of Snorkel. The LFs are user-defined and can take many forms, including heuristics, rule-based systems, and external knowledge bases. Here we define the LFs to consider the semantic similarity between embeddings of category definitions and abstracts. LF definition is the only user input required for generating programmatic labels with Snorkel.

Second, given that LFs are typically imprecise and noisy, numerous labeling functions are combined in a generative model, which is automatically learned, to produce a final set of labels in Snorkel. Snorkel's generative models are probabilistic graphical models that allow users to express their uncertainty in labeling functions and capture label dependencies. Instead of using ground truth data, this step learns from the agreement and disagreement among the labeling functions and produces probabilistic labels. 

Third, these labels are used to train a discriminative classification model, which is a supervised learning model that takes the programmatic labels produced by the generative model as input and learns to predict the labels of the new instances. The discriminative model generalizes beyond the logic contained in the LFs to achieve better classification performance on new data than would be achieved by solely relying on the developed generative model.

\begin{figure}[!ht]
\centering
   \includegraphics[width=\linewidth]{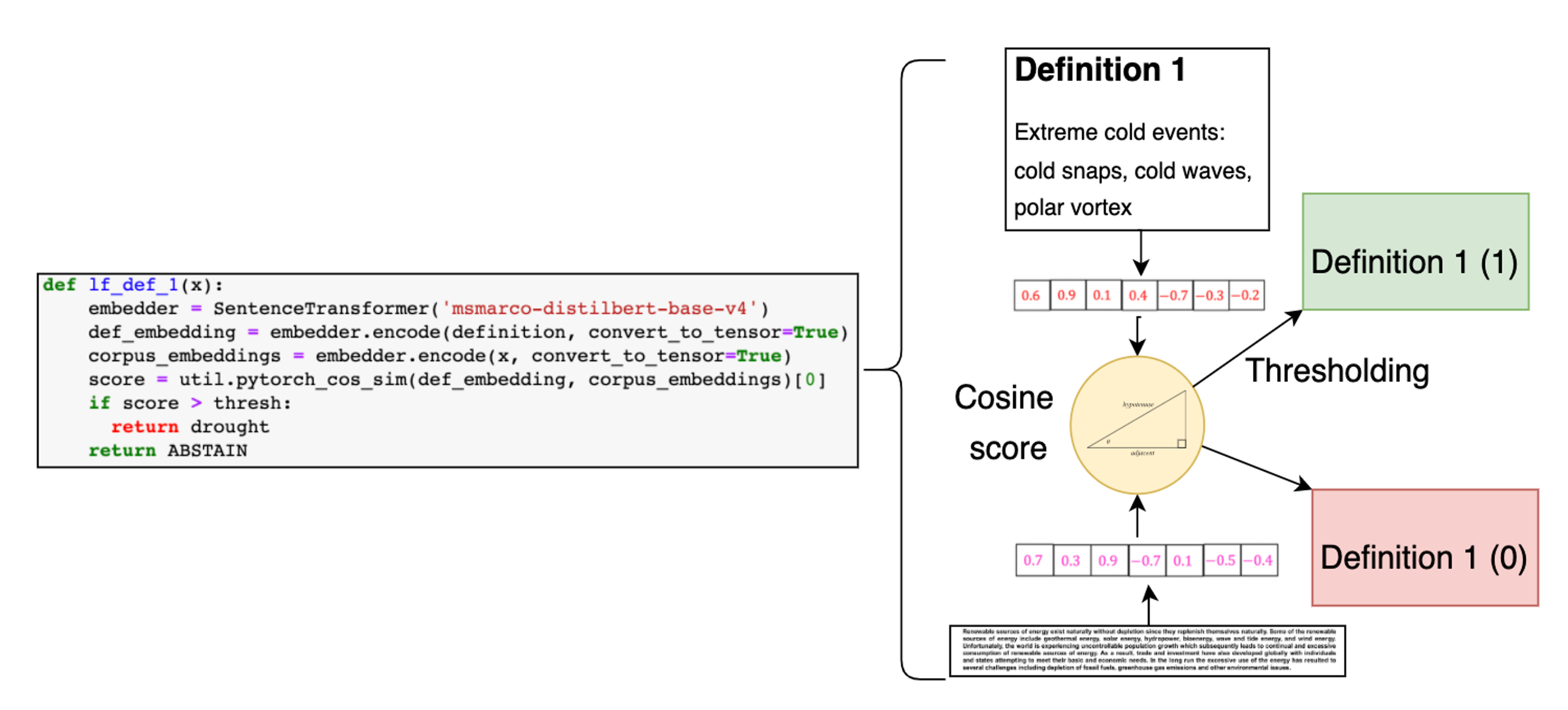}  
\caption{Labeling function (LF)} 
\label{fig_lf}
\end{figure}

\subsubsection{Defining the LFs using multiple embeddings}\label{sec_mult_embedd}
Defining the labeling function is the first and most important step in the weak supervision using Snorkel.
The  common method for defining LFs for NLP is to use keyword searches to find specific words in the documents or pattern matching to find specific syntactical patterns. Often, semantically similar documents do not share the same keywords or patterns. Moreover, finding representative keywords or patterns is difficult, and the number of LFs required to cover all possible keywords or patterns will be enormous. As a result, these approaches are ineffective for writing LFs and may require nearly as much contribution from subject-matter experts as do conventional manual labeling processes.

In this paper we propose a novel approach to defining LFs based on semantic similarities between climate categories and documents. Each of our climate hazard and NCF categories has a definition. A document belongs to a specific category if the document and the definition of climate or NCF are semantically similar. To determine semantic similarity, we first embed the document and definitions using a sentence transformer model \cite{reimers-2019-sentence-bert}, which represents the text as a numeric vector. A sentence transformer is a framework that includes pretrained models for calculating dense vector representations of sentences. Semantically similar sentences have similar representations in the embedding space. The models are built on transformer networks such as BERT \cite{devlin2018bert} and RoBERTa \cite{liu2019roberta}, to achieve cutting-edge performance in a variety of tasks. Sentence transformers offer a number of models trained on millions of diverse training datasets such as S2ORC citation pairs (the open-source corpus we used to build our climate corpus), Stack Exchange title and body pairs, WikiAnswers duplicate question pairs, queries from the Bing search engine, and Google queries and Google featured snippets. Each of these models is fine-tuned for a specific purpose. As a result, defining a labeling function based on a single embedding technique may be insufficient, resulting in poor labeling performance. To overcome this issue, we introduce a novel technique to define a generative model comprised of multiple LFs using different embedding techniques. 

After embedding the category definitions and documents, a similarity measure such as cosine similarity, Euclidean distance, or dot product can be used to evaluate the semantic similarity of each document to each category definition. The similarity score between the embeddings of a category definition and a document is compared with a threshold value to determine whether or not the document belongs to the category. Figure \ref{fig_lf} depicts how the labeling function is defined by comparing the cosine similarity between the embeddings of documents and the definitions to a threshold value.

In our proposed approach, each LF uses a different embedding technique to embed every category definition and document in the climate and NCF corpus. Next, each LF computes the cosine similarity between the embeddings of each category definition and each document. Finally, each LF compares each cosine similarity score to determine whether or not a document fits into a category. The generative model's parameters are set to update weights and combine the labels produced by the LFs. To determine the labels, the model assumes that the LFs are conditionally independent. 

Since the input documents in this case are multi-label in nature, a single document may fall under more than one category. Snorkel does not provide an explicit function to perform multi-label labeling. Hence, to perform multi-label labeling, we label the documents for each of the categories independently, resulting in $N$ Snorkel generative models built to perform labeling for $N$ categories (i.e., one generative model per category). Each model performs binary labeling to determine whether or not a document belongs to a category. The final set of labels for each document is the union of all labels for the document determined from the $N$ Snorkel models.

\begin{figure}[!ht]
\centering
   \includegraphics[width=\linewidth]{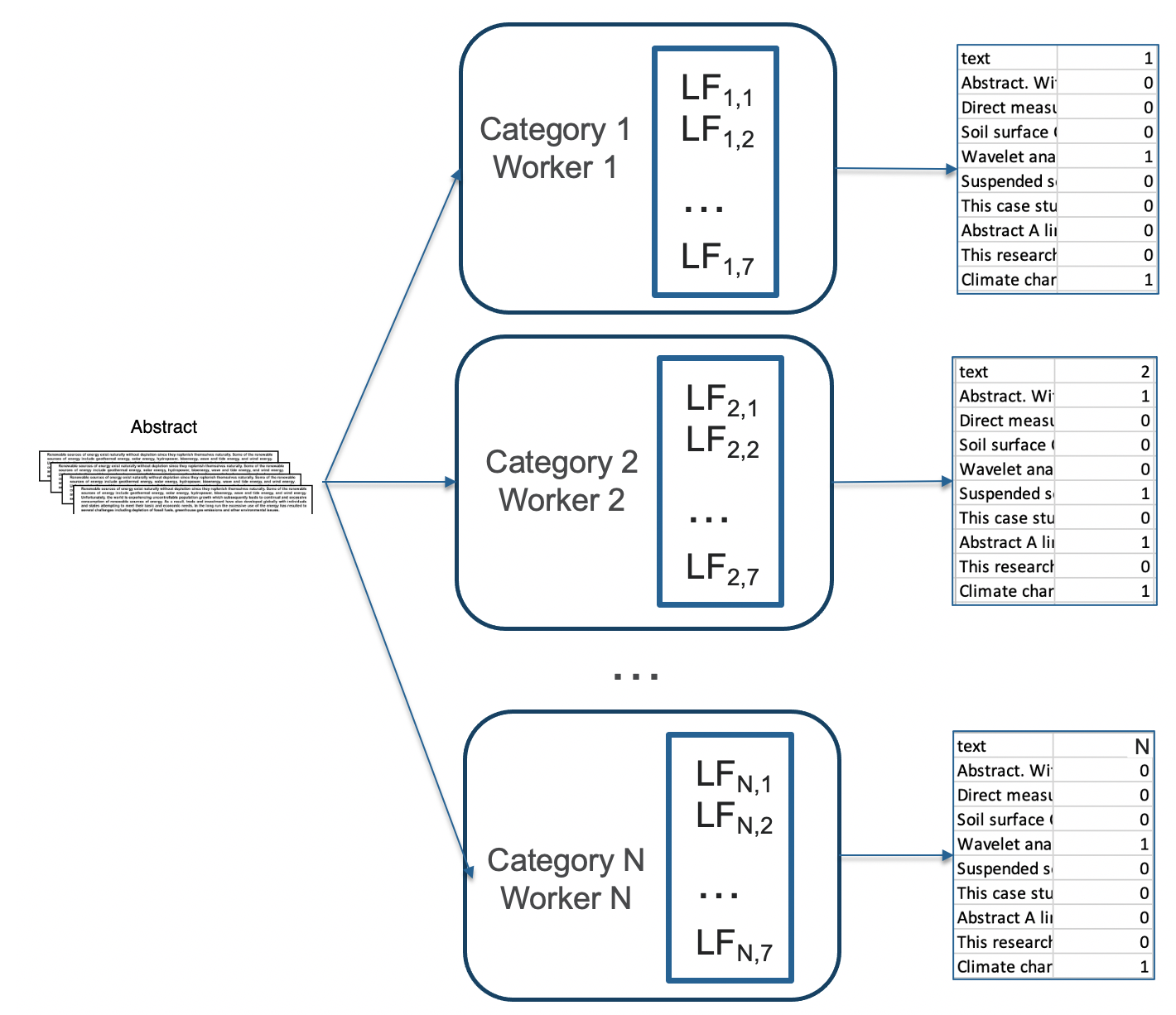}  
\caption{Scaling weak supervision on GPU cluster} 
\label{fig_scaling}
\end{figure}

\subsection{Scaling Weak Supervision}
Building multiple models to label multiple categories is computationally expensive and time-consuming if the task is performed sequentially. Hence, we scale the Snorkel 
training over multiple workers as shown in Figure \ref{fig_scaling}. Independent Snorkel models are trained for $N$ categories simultaneously on $N$ workers. Each Snorkel model receives the same unlabeled documents. Each worker is responsible for labeling each category, and independent models run on each worker without interaction. Simultaneous execution across $N$ workers reduced the computational time by a factor of roughly $N$, allowing us to label a large number of documents in a short period of time. 
We combine the labels produced by each Snorkel model across multiple GPUs to create our training set for the supervised learning model.


\begin{figure*}[!t]
\centering
   \includegraphics[width=\linewidth]{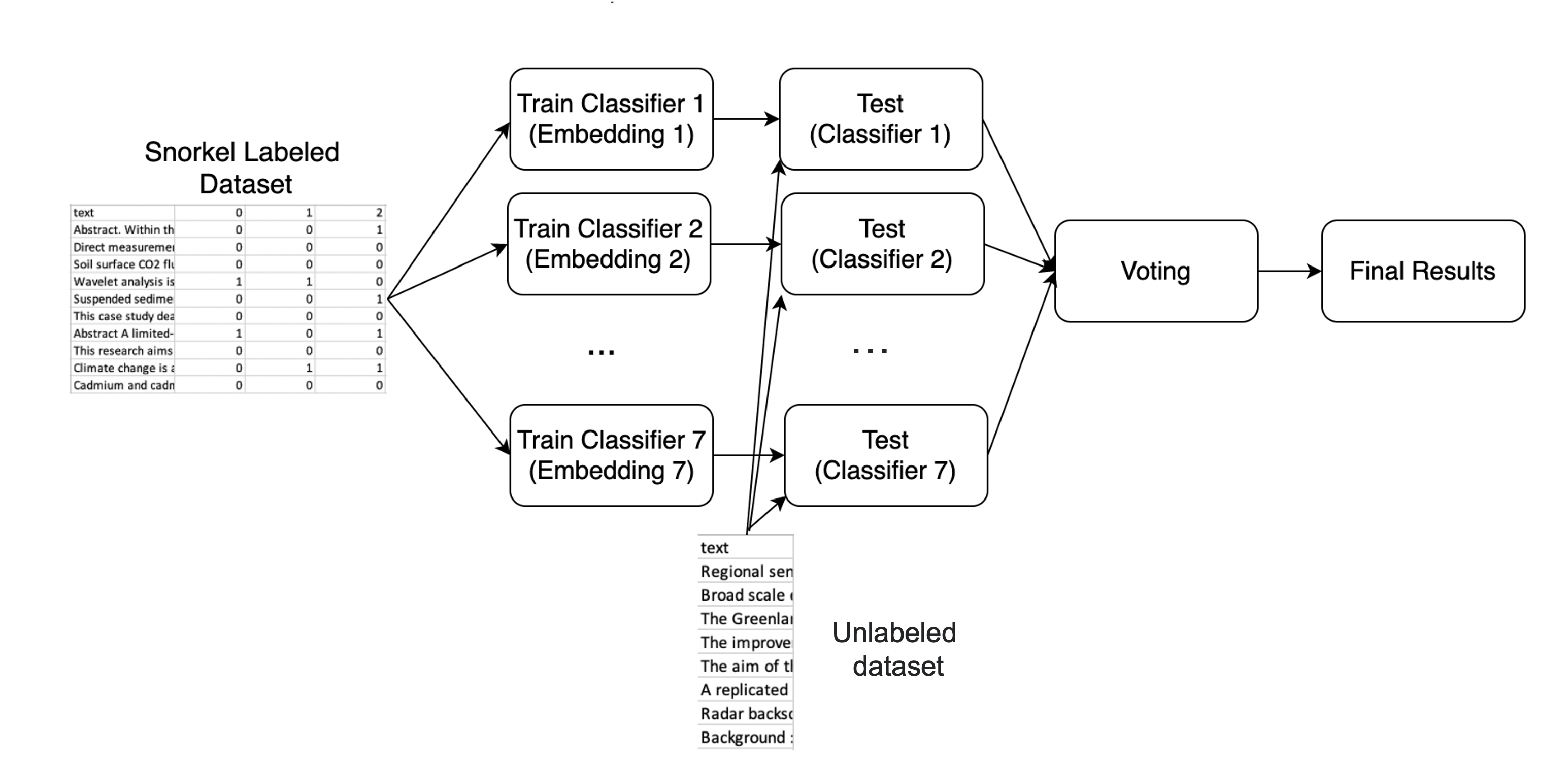}  
\caption{Supervised learning using Snorkel labeled dataset} 
\label{fig_supervised}
\end{figure*}

\subsection{Supervised Learning Using Labeled Dataset} \label{sec_sl}
Given the Snorkel-labeled training dataset, we train a supervised learning model to generate the labels for the rest of our unlabeled dataset. The goal is to develop a discriminative model that generalizes beyond the training dataset. Here we perform multilabel classification, where one document can belong to multiple categories. The most common approach for multilabel classification is binary relevance \cite{read2011classifier} where the multilabel problem is transformed into multiple single-label or binary classification problems. In this way, multiple single-label classifications are made by using a single-label classifier, and these classification results are subsequently combined into multilabel representations. The basic assumption of the binary relevance method is that the multiple labels are independent of each other. Because of this transformation process, the method ignores the inherent correlation that exists between the labels. Thus, binary relevance would be unable to capture the known dependencies across several climate change hazard categories and across NCF categories.  In this paper we use classifier chains \cite{read2011classifier} for our classification task. 
The classifiers in a classifier chain are set up in a chain so that the output of one classifier feeds into the input of the next, and so forth. Using the initial feature set as input, the first classifier in the chain is trained to predict the first label in the dataset. This classifier's output is then appended to the second classifier's feature set as a new feature, and the second classifier is trained to predict the second label using the expanded feature set as input. A chain of classifiers is built by repeating this procedure for each label in the dataset. By taking into account the interdependencies between labels, a classifier chain  improves the performance of multilabel classification.

The supervised learning pipeline is shown in Figure \ref{fig_supervised}. Instead of using a single classifier, we use an ensemble of classifiers using multiple embedding techniques.  The Snorkel-labeled dataset is used to train  multiple classifiers chain models. For each classifier, the text is converted to a numerical vector using a different BERT-based embedding model as discussed  in Section \ref{sec_mult_embedd}. The numerical vector representation is used to train the classifiers. After training, the model predicts multiclass labels for the unlabeled subset of the corpus.
The same embedding techniques are used to convert the unlabeled dataset into a numeric vector. The  classifiers then produce the labels for the unlabeled dataset. The majority voting method is used to determine the final label based on the results of multiple classifiers. A document receives a label 1 or 0 if more than 50\% of the classifiers are producing the same findings. The ensemble classification model offers improved classification performance over a classifier built on a single embedding technique. 

Overall, the pipeline we have shown makes it possible to label huge datasets quickly and without the need for human interaction. The labeled dataset enables us to perform unsupervised topic modeling on the pair of  climate hazards and NCF categories in order to retrieve insightful data.

\subsection{Category-wise Topic Modeling}
The labeling of the entire corpus enables us to conduct category-based topic modeling to comprehend emerging topics focusing on the impact of climate change on critical infrastructure.
Topic modeling is an effective unsupervised technique for identifying abstract topics present in a set of documents. Latent Dirichlet Allocation  \cite{blei2003latent} is the most commonly used model, which describes a document as a bag-of-words and represents it as a mix of latent topics. The main problem with LDA is that it ignores the semantic similarity of the words. 
Therefore, we adopted BerTopic  \cite{grootendorst2022bertopic}, which leverages the BERT embedding technique to exploit the semantic similarity of the documents. 

To obtain document-level information, BerTopic first creates document embeddings with pretrained BERT-based models. Thereafter, the dimensionality of document embeddings is reduced using Uniform Manifold Approximation and Projection for Dimension Reduction  \cite{mcinnes2018umap}. The reduced embeddings are clustered by using hierarchical density-based spatial clustering or HDBSCAN \cite{mcinnes2017hdbscan} to form semantically similar clusters of documents. Each of these clusters  represents a distinct topic. Next, a cluster-based term frequency-inverse document frequency (TF-IDF) or c-TF-IDF is developed to find the importance of words in clusters instead of individual documents. The TF-IDF is transformed to take into account the cluster frequency instead of the document frequency as follows:  
\begin{equation*}
    W_{t,c} = tf_{t,c} \cdot  log(1+ \frac{A}{tf_t}),
\end{equation*}
where the term frequency $tf_{t,c}$ represents the frequency of term $t$ in cluster $c$. Cluster $c$ is a collection of documents concatenated into a single document. The inverse document frequency or IDF is replaced by inverse cluster frequency. This measures the importance of a word in a cluster instead of a document. It is determined by taking the logarithm of the average number of words in each cluster $A$ divided by the frequency of the term $t$ across all clusters.
The topic-word distributions for each group of documents are produced by using c-TF-IDF.


%% file: results.tex

Here we assess the performance of each phase of the proposed method, including the building of the climate and infrastructure corpus,  weak supervision, supervised learning, and category-wise topic modeling.  We discuss the advantages and disadvantages of our proposed methodology, as well as the capabilities it provides for analyzing the impact of climate change on critical infrastructure. We demonstrate that our proposed method is more effective in finding relevant information by comparing the outcomes with ChatGPT, a large language model chatbot developed by OpenAI. 

The experiment was carried out on a GPU-based cluster. This cluster contains 126 nodes. Each node is equipped with two 2.4 GHz Intel Haswell E5-2620 v3 CPUs and one NVIDIA Tesla K80 dual-GPU card (two K40 GPUs).


\subsection{Building a Corpus of Climate and Infrastructure Data} \label{sec_res_pre}

The first step in our proposed approach is to separate climate and NCFs-related documents from the entire corpus of 76M documents. We used a rule-based classifier to classify the documents that are similar to the category definitions. To perform the rule-based classification, we computed the embedding of the documents and definitions using the multi--qa--mpnet--base--cos--v1 sentence transformer's  \cite{reimers-2020-multilingual-sentence-bert} pretrained model. The model was pretrained on 215 million question-and-answer pairs from various sources, and it mapped documents to a 768-dimensional dense vector space. 
Then, as described in Section \ref{sec_preproc}, we performed rule-based  classification and down-select the climate and NCFs corpus from the entire S2ORC corpus. Through experimentation, we discovered that documents with a maximum similarity score across climate hazard definitions of (i) 0.2 or lower are unrelated to climate, (ii) 0.2--0.4 are a mix of climate-related and unrelated documents, and (iii) 0.4 or higher are very likely related to climate. Therefore, we select the threshold of 0.4 to filter climate-related documents from the entire corpus. 
Running the preprocessing on a single machine for 76M documents is computationally expensive. 
Therefore, we utilized 100 GPUs from the  GPU cluster 
to run the preprocessing operation. Preprocessing takes place in 10 hours, and each GPU processes the 76M documents. We then filtered out 604,621 climate-related documents from the corpus of 76M documents. 

Using the same methods mentioned above, we looked for NCF-related materials among the 604,621 climate-related documents. Using the same sentence transformer model 
, we searched for similarities between the 55 NCF definitions and 604,621 climate-related documents. We used 100 GPUs from the  GPU cluster, each of which processes approximately 6,000 documents within 10 minutes. 
We obtained 17,136 NCFs and climate-related documents.


\subsection{Weak Supervision}

After obtaining the climate and NCF-related corpus, we labeled a portion of the dataset using the proposed methodology of weak supervision employing multiple embedding techniques. We labeled part of the corpus for the 18 climate hazard categories and 55 NCF categories to generate a training dataset for supervised learning. This solution overcomes the data labeling bottleneck by labeling a large number of data points automatically without the need for human participation. We used the Snorkel system \cite{Ratner2022} to rapidly generate the training data.


Our novel methodology, described in Section \ref{sec_ws}, used seven different embedding techniques to build the LFs for each category. 
The embedding techniques are listed in Table \ref{tab_embeddings}. 
Each of the LFs uses one of the embedding techniques to calculate the similarity score between the document and the definition. If the score exceeds a certain threshold, the LFs indicates that the document falls into that category.  Snorkel then combines the results of all LFs while minimizing overlaps and conflicts. Here we used the same threshold of 0.4 as mentioned earlier in the preprocessing steps (Section \ref{sec_res_pre}). 

\begin{table}[!ht]\centering
\begin{tabular}{|l|l|}
\hline
\textbf{Embeddings}            & \textbf{Dimensions}  \\ \hline
multi-qa-distilbert-cos-v1     &    768                                        \\ \hline
multi-qa-MiniLM-L6-cos-v1      &    384                                           \\ \hline
bert-base-nli-stsb-mean-tokens &    768                                        \\ \hline
all-mpnet-base-v2              &    768                                       \\ \hline
all-distilroberta-v1           &    768                                         \\ \hline
multi-qa-mpnet-base-cos-v1     &    768                                        \\ \hline
msmarco-distilbert-base-v4     &    768                                          \\ \hline
\end{tabular}
\caption{Different embedding techniques used for weak supervision}
\label{tab_embeddings}
\end{table}

Using Snorkel, we label 5,383 documents in order to obtain sufficient training data for all categories. 
On a single GPU, labeling a single category for 5,383 documents takes approximately 12 hours. 
It would have taken 12 * 18 = 216 hours, or 9 days, to label 18 climate categories.  To reduce the computational time, we scaled weak supervision across multiple GPUs. We assigned different GPUs to label different categories because they are labeled independently. We ran weak supervision on 18 GPUs for 18 climate categories. It takes 12 hours to label 5,383 documents simultaneously.

Following the same process used for labeling a subset of the corpus with climate hazard labels, we label 5,000 documents with NCF labels using weak supervision in order to obtain sufficient training data for all categories. Each of the 55 categories is labeled separately. Because each category is labeled independently, we allocate distinct GPUs to each one. On 55 GPUs, we execute weak supervision for 55 NCF categories. Labeling 5,000 documents simultaneously took approximately 10 hours.

\subsection{Supervised learning}
We use the Snorkel-labeled dataset for training our supervised learning model. Since our dataset is multilabel,  we use a classifier chain to perform multilabel classification as discussed in Section \ref{sec_sl}.

\begin{figure}[!ht]
\centering
   \includegraphics[width=\linewidth]{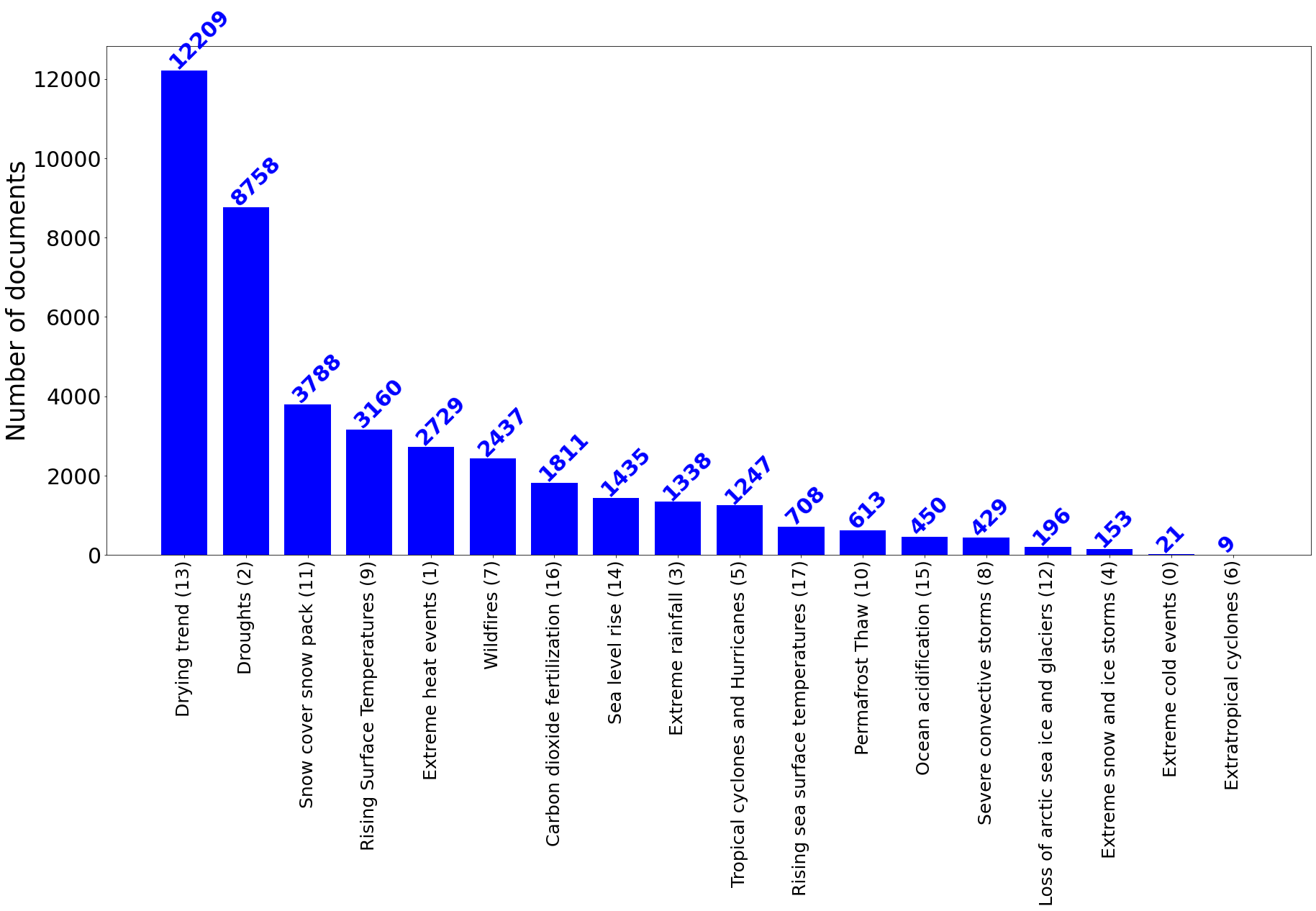}  
\caption{Document per class for climate hazards} 
\label{fig_per_class_climate}
\end{figure}

\begin{figure}[!ht]
\centering
   \includegraphics[width=\linewidth]{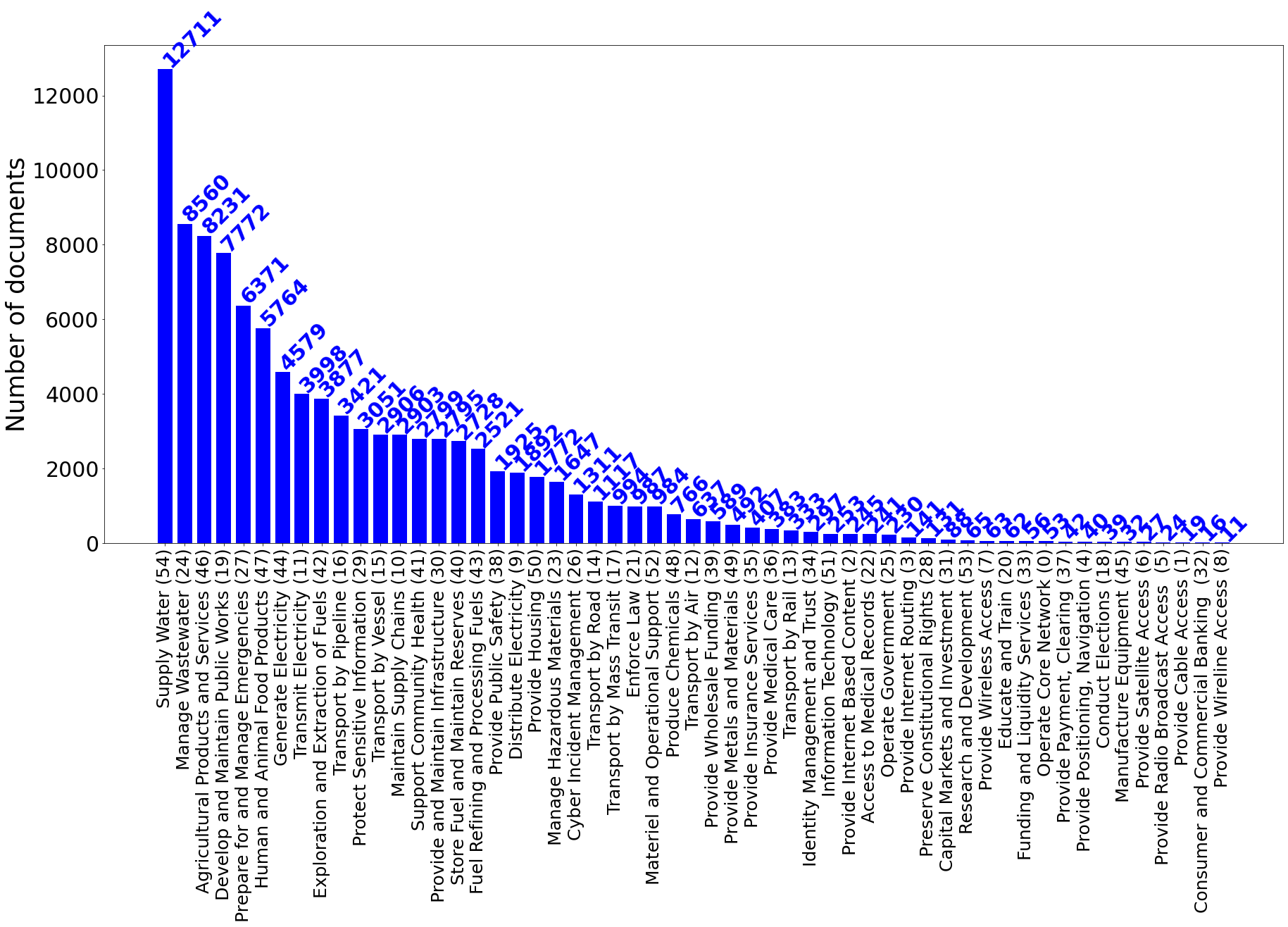}  
\caption{Document per class for NCFs} 
\label{fig_per_clas_ncf}
\end{figure}

Prior to training the supervised learning model on the entire training set, we performed 10-fold cross-validation on the Snorkel-labeled dataset to obtain an estimate of the predictive performance of our supervised model once trained. We divided 5,383 climate-related documents into ten consecutive folds. Each fold was validated once after training the supervised learning model with the other nine folds. We achieve 91\% accuracy using 10-fold cross-validation. 

Next, we trained the classifier chain model for climate hazard categories using the 5,383 documents in the climate training set (i.e., with assigned multiclass climate hazard category labels) and predict multiclass labels  for the remaining 11,753 unlabeled documents of the corpus. We trained the classifier with balance class weight to adjust the weight of the minority class to make it equal to the weight of the majority class.
We trained seven classifiers using seven different embedding techniques and performed majority voting to find the final label as shown in Figure \ref{fig_supervised}. The seven classifiers are trained simultaneously across seven GPUs. It takes less than one hour to train and test on each GPU.  
After predicting the climate labels of 17,136 documents, we plotted the number of documents per climate hazard category to explore the prevalence of the individual climate hazard categories (Figure \ref{fig_per_class_climate}). 
The results show that Drying Trend is the most discussed topic as 12,209 documents belong to this category. The plot demonstrates how most of the topics that receive a lot of attention are related to Drying Trends, Drought, Rising Surface Temperature, and their consequences, including Wildfire and Extreme Heat Events. Four of the top five most highly discussed topics are associated with increasing air temperatures: Drying Trend, Drought, Rising Surface Temperature, and Extreme Heat Events. This result is intuitive because concerns surrounding global warming have been widely published over the past few decades.
Extreme Snow and Ice Storms, Extra-tropical Cyclones, and Cold Events are discussed the fewest times in our climate corpus. 

Following the same process used for developing the  supervised model for the climate hazard, we trained the classifier chain model for NCF categories using the 5,000 documents with NCF labels and used the trained model to predict classes for the remaining 12,136 documents. The seven classifiers were trained and tested simultaneously across seven GPUs and produced the results in less than one hour. We trained the classifier with a balanced class weight to make the weight of the minority class equal to the weight of the majority class.

After predicting the NCF labels of 17,136 documents, we plotted the number of documents per NCF category in Figure \ref{fig_per_clas_ncf}. The results show that Supply Water is the most discussed topic as 12,711 documents belong to this category. 
The most common NCF categories in our corpus are Supply Water, Wastewater Management, and Agricultural Products and Services; this result is logical given that much research has been conducted to assess the potential impacts of climate change on operations in these topics. On the other end of the spectrum, Provide Wire-line Access, Consumer and Commercial Banking, and Provide Cable Access are the least discussed NCFs across the corpus, indicating less historical research on the potential impacts of climate change on these NCFs.


\subsection{Topic Modeling on Climate Hazards and NCFs Pairs}
\begin{figure*}[!ht]
\centering
   \includegraphics[width=\linewidth]{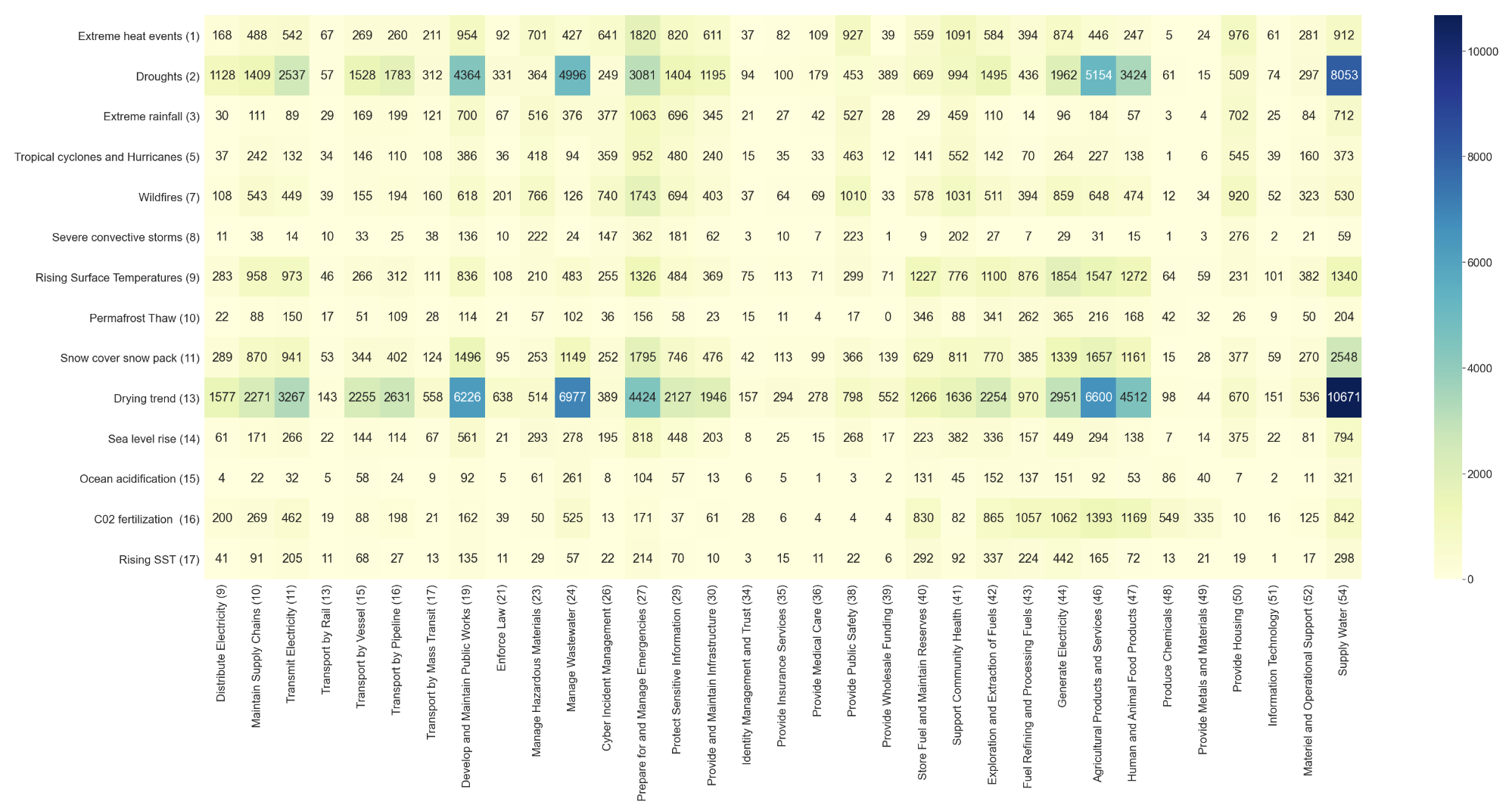}  
\caption{Number of documents belonging to each climate hazard and NCFs pair} 
\label{fig_corr}
\end{figure*}

After labeling the dataset with climate hazard and NCF labels, we investigated the overlap between document label pairs. For this effort, we counted the number of documents associated with each NCF and each climate hazard category. To explore the prevalence of climate hazard categories across NCF categories, we plotted counts of documents associated with each climate hazard and NCF pair (Figure\ref{fig_corr}). To improve the readability of this plot, we excluded any climate hazard category (row) or NCF category (column) if less than 100 documents are associated with every climate hazard and NCF category (cells) in that row or column. In this figure, darker colors indicate that more documents in the corpus have both row and column labels. The pair with the most documents is Drying Trend and Supply Water, which has 10,671 documents. Identifying the prevalence of intersecting pairs of climate hazards and NCFs expedites the analytical process required to understand climate risks to critical infrastructure and to inform targeted and prioritized adaptation strategies.

To identify actionable information on climate hazards and threats to critical infrastructure, we performed topic modeling on the climate and NCF pairs. 
We used BERTopic to perform topic modeling on the pairs and saved the model for future visualization. We preprocessed the data and removed stopwords (a popular NLP preprocessing step used to remove the most common, and least informative, words from text such as ``and" or ``the") before performing topic modeling. The preprocessed data is fed into topic modeling, and the model returns topics and the associated count of documents in the climate hazard/NCF pair subset assigned to each topic. 
The resulting topics suggest different aspects (e.g., causal factors) and complexities (e.g., upstream and downstream dependencies) that should be considered by decision-makers when trying to understand and address the challenges surrounding the climate hazard and NCFs intersection. Further, while some of the topics point to well-established concepts, others seem to indicate more obscure topics that may not be easily identified by non-experts. To illustrate the results of the topic modeling approach, we next present two case studies for specific climate hazard and NCF pairs.

\subsubsection{Case Study I: Topics at the Intersection of Drying Trend and Supply Water}
\begin{figure}[!ht]
\centering
   \includegraphics[width=\linewidth]{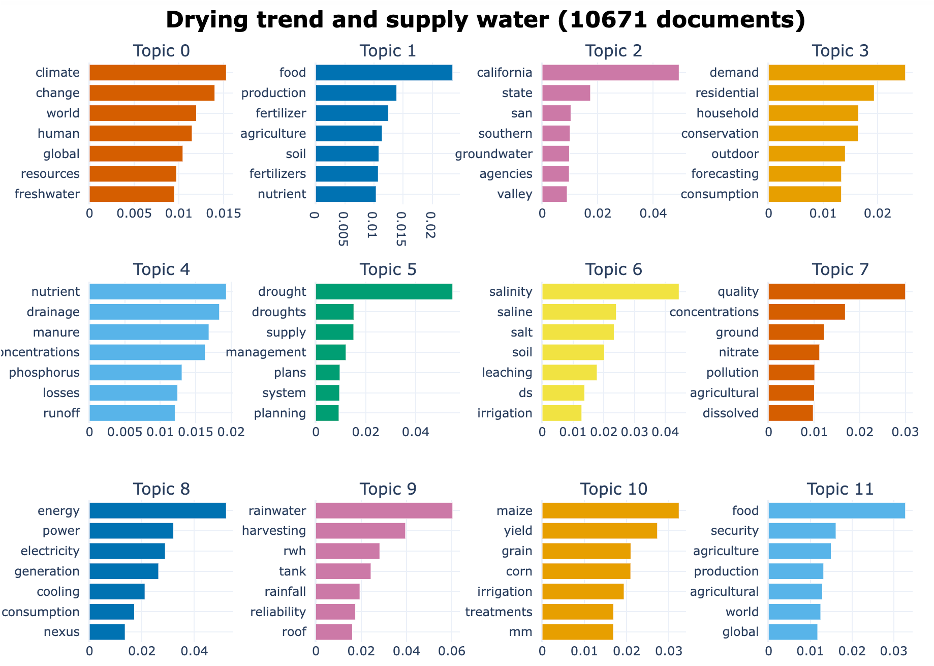}  
\caption{Topic modeling for drying trend and supply water} 
\label{fig_tm_dt}
\end{figure}

We executed topic modeling on the set of 10,671 documents with both the Drying Trend climatic hazard and Supply Water NCF labels.
In Figures \ref{fig_tm_dt} and \ref{fig_tm_er}, topics are shown in descending frequency (i.e., Topic 0 is the most frequently occurring topic in each figure as a maximum number of documents belongs to Topic 0). The topic modeling results for the Drying Trend climate hazard and Supply Water NCF combination identify several topics of interest, including topics that may indicate correlation or dependency relationships with other NCFs or climate hazards. For instance, the second most frequent topic (topic 1) indicates significant discussion in the literature surrounding drying trends and supplying water in the context of agriculture, indicating potential impacts on the Food and Agriculture sector that produces and provides agricultural products related to other  NCFs. In fact, agriculture aspects can be seen across several topics, including topics 7, 10 and 11. Similarly, topic 8 indicates that the intersection of Drying Trends and Supply Water may be particularly relevant to the Generate Electricity NCF and the Energy sector, particularly the electric power subsector. Topic 5 suggests that the intersection of supply water and drying trends may have some correlation or dependency relationship with the drought climate hazard. Topic 2 identifies particular geographies where the intersection of drying trends and the Supply Water NCF may be particularly important. Some of the topics also seem to indicate correlation or upstream or downstream dependencies with other NCFs and climate hazards. 

To learn more about the various aspects discussed in the topics and to assess the performance of the weak supervision process, we chose two topics for each case study and manually reviewed the top five abstracts (i.e., the abstracts with the highest probability for the particular topic) for each chosen topic.
We aimed to select an expected topic and a surprising or unexpected topic. From Figure \ref{fig_tm_dt}, we selected topics 2 and 8. Eight of the ten abstracts reviewed for the two selected topics for the supply water NCF and drying trend climate hazard were deemed relevant to at least one of the two labels, while two were determined to be less relevant to either label. 
The five abstracts reviewed from topic 2 in Figure \ref{fig_tm_dt} discuss challenges associated with managing water supplies in southern California in the face of drying trends and historic droughts, including  the role of water storage and water rights policies and strategies, as well as the role of legislation in water supply management. From topic 8 in Figure \ref{fig_tm_dt}, the five abstracts reviewed discuss considerations of water consumption and impacts of water availability on electric power generation technologies. Overall, the topic model performs well, identifying diverse and informative aspects from the documents. Furthermore, the manual review of abstracts in support of this case study suggests that the labels generated by weak supervision and supervised learning models are achieving satisfactory performance.

\subsubsection{Case study II: Topics at the Intersection of Extreme Rainfall with Inland and Coastal Flooding and Prepare for and Manage Emergencies}

\begin{figure}[!ht]
\centering
   \includegraphics[width=\linewidth]{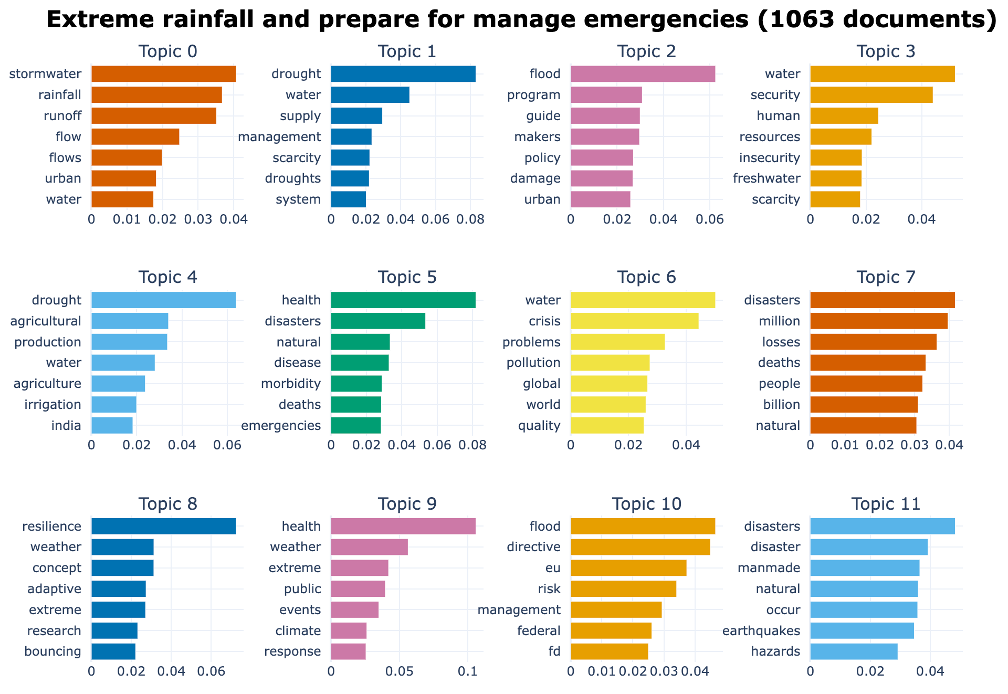}  
\caption{Topic modeling for Extreme Rainfall and Prepare for and Manage Emergencies} 
\label{fig_tm_er}
\end{figure}

Figure \ref{fig_tm_er} shows topics identified in articles with both the Extreme Rainfall climate hazard label and  the Prepare for and Manage Emergencies NCF label. Several topics appear to suggest commonalities with other NCFs, including manage wastewater (topic 0), provide water (topics 1, 3, and 6), provide medical care (topics 5 and 9), and produce and provide agricultural products and services (topic 4). Similarly, drought (topics 1 and 4) is identified as an additional climate hazard relevant to the intersection of Extreme Rainfall and Prepare for and Manage Emergencies. Topics also span across geographic scales, including city-level (topics 0 and 2), national (topic 4), continental (topic 10), and global (topic 6). At the city-level scale, topics 0 and 2 address urban rainfall concerns (runoff, flows, storm water, damage), while at the global scale, topic 6 addresses issues surrounding water quality and pollution at the global scale. A few topics discuss flood management and policies (topics 2 and 10).

To explore the performance of the topic modeling, we selected topics 0 and 1 from Figure \ref{fig_tm_er} and manually reviewed ten abstracts.  For the two selected topics, nine of the ten abstracts reviewed were deemed relevant to at least one of the two labels. 
The results suggest the favorable performance of the weak supervision and/or trained supervised learning model. In Figure \ref{fig_tm_er}, the five abstracts reviewed from topic 0 discuss issues surrounding stormwater management, including issues of pollutant transport and water quality issues in receiving bodies of water; and the five abstracts reviewed from topic 1 discuss the role of policy- and decision-making in water supply management to avoid future disasters.  This manual review is performed to confirm the performance of weak supervision. However, we may be able to increase the effectiveness of weak supervision even further by adjusting the threshold for each category to reduce the number of irrelevant papers in each topic.

Overall, topic modeling on climate hazard and NCF pairs reveals different impacts of climate change on critical infrastructure. It also supports more targeted manual review of documents by identifying major topical themes across the corpus subset relevant to the NCF and climate hazard combination, enabling users to pursue documents associated with topics of interest. For example, by studying the topics identified for the drying trend and supply water pair, an emergency manager may want to conduct a risk assessment focused on electricity generation reliability (topic 8). Similarly, an engineer can use this information to better identify design considerations (e.g., changing environmental conditions) for future electric power generation technologies. Ultimately, the individual using the results of the topic modeling would be able to focus their limited time and resources on a smaller set of relevant papers to extract pertinent information on a topic of interest and inform analyses and decision-making.


\subsection{Comparison with ChatGPT}
Here we compare the outcomes of our proposed approach with ChatGPT, a generative pretraining transformer that can conduct conversations on a variety of subjects. We carried out experiments to understand the response of ChatGPT compared with our proposed method. We asked the question to ChatGPT  \textit{What is the impact of drying trend on Supply Water?}. We received the following answer:

``A drying trend can have a significant impact on the availability of water for domestic, industrial, and agricultural use. Drought conditions can result in a reduced supply of surface water, such as lakes and rivers, as well as a decline in ground water levels. This can lead to water shortages and rationing, particularly in areas that rely heavily on surface water or ground water for their water supply. In addition, a drying trend can also impact the treatment and distribution of water. During drought conditions, the quality of surface water may decline due to low flows and increased concentration of contaminants. This can lead to an increased demand for treated water, which may put additional strain on water treatment facilities. Overall, the impact of a drying trend on the supply of water can be significant and can have a range of consequences, including reduced water availability, higher water costs, and increased demand for treated water. It is important to carefully monitor and manage water resources to ensure that they are available for use and to mitigate the impact of a drying trend on the water supply.'' 

There are no particular insights in this answer, which is very general. In contrast, our topic modeling on Drying Trend and Supply Water, as shown in  Figure \ref{fig_tm_dt}, revealed very interesting insights about specific areas such as the impact of drying trends and supply water on the agriculture sector, California groundwater,  drought adaptation planning, and electric power sector. 

We posed several similar questions to ChatGPT and received the same generic response. ChatGPT performs remarkably well when offering general solutions as a result of its extensive training on the generic corpus. However, it omits some crucial information that we are interested in. It does not uncover all of the essential information about the relationship between climate hazards and NCFs when compared with the results of our topic modeling. ChatGPT also does not position users to explore a curated subset of relevant research papers that can help them evaluate adaptation strategies and inform decisions.




%% file: conclusion.tex
In this paper, we described 
a novel methodology for labeling and analyzing a large corpus of documents in a short amount of time using weak supervision. We developed the weak supervision pipeline using Snorkel and employed multiple embedding techniques to improve the labeling of documents compared with labeling documents using a single embedding. We performed multilabel labeling by combining multiple binary labeling models. The subset of the corpus labeled by using weak supervision was used as a training dataset for building a supervised learning model, which supports large scaling and generalization beyond the training dataset. We scaled weak supervision and supervised learning across several workers to reduce the overall labeling and classification times. Our method labeled the entire set of documents across 18 climate change hazards and 55 NCF categories in approximately 13 hours, which represents a significant reduction from the time necessary through conventional manual labeling of the data by subject-matter experts. 
Unlike active learning and manual labeling techniques, our proposed method does not require human intervention, and the proposed method is not restricted to our current corpus. Our method can label any corpus of documents given category definitions. Furthermore, the categorization helps us perform category-based topic modeling and identify overarching topics in the literature for specific climate hazards, NCFs, and combinations of climate hazards and NCFs. 
Category-based topic modeling is important for understanding and addressing  multifaceted relationships between different factors such as drying trends and water supply. It allows scientists and policymakers to target a very small subset of documents for analysis in order to understand the relationships between different topics and to predict the potential impacts of different scenarios. Thus, our proposed methodology can help decision-makers gain a better understanding of the topic of interest and to make informed decisions based on the best available evidence.


%% file: appendix.tex
\subsection{Climate Hazards}\label{sec_climate}
Here are the 18 climate hazards:
(1) Extreme cold events,  
(2) Extreme heat events,
(3) Droughts, 
(4) Extreme rainfall, 
(5) Extreme snow and ice storms,
6) Tropical cyclones and hurricanes,
(7) Extratropical cyclones,
(8) Wildfires,
(9) Severe convective storms,
(10) Rising surface temperatures,
(11) Permafrost thaw,
(12) Snow cover snow pack,
(13) Loss of arctic sea ice and glaciers,
(14) Drying trend,
(15) Sea-level rise,
(16) Ocean acidification,
(17) Carbon dioxide fertilization,  and
(18) Rising sea surface temperatures.

\subsection{National Critical Functions Set}\label{sec_ncfs}
Here are 55 \href{https://www.cisa.gov/national-critical-functions-set}{National Critical Functions} under four (connect,  distribute, manage, supply) broad categories.

The NCFs under connect categories are as follows:
(1) Operate Core Network, (2) Provide Cable Access Network Services, (3) Provide Internet Based Content, Information, and Communication Services, (4) Provide Internet Routing, Access, and Connection Services, (5) Provide Positioning, Navigation, and Timing Services, (6) Provide Radio Broadcast Access Network Services, (7) Provide Satellite Access Network Services, (8) Provide Wireless Access Network Services, and (9) Provide Wireline Access Network Services.
The NCFs under distribute categories are (10) Distribute Electricity,
(11) Maintain Supply Chains, (12) Transmit Electricity, (13) Transport Cargo and Passengers by Air, (14) Transport Cargo and Passengers by Rail, (15) Transport Cargo and Passengers by Road, ((18) Transport Passengers by Mass Transit. 
The NCFs under manage categories are as follows:
(19) Conduct Elections, (20) Develop and Maintain Public Works and Services, (21) Educate and Train, (22) Enforce Law, (23) Maintain Access to Medical Records,
(24) Manage Hazardous Materials, (25) Manage Wastewater, (26) Operate Government,
(27) Perform Cyber Incident Management Capabilities, (28) Prepare for and Manage Emergencies, (29) Preserve Constitutional Rights, (30) Protect Sensitive Information, (31) Provide and Maintain Infrastructure, (32) Provide Capital Markets and Investment Activities, (33) Provide Consumer and Commercial Banking Services, (34) Provide Funding and Liquidity Services, (35) Provide Identity Management and Associated Trust Support Services, (36) Provide Insurance Services,
(37) Provide Medical Care, (38) Provide Payment, Clearing, and Settlement Services,
(39) Provide Public Safety, (40) Provide Wholesale Funding, (41) Store Fuel and Maintain Reserves, and (42) Support Community Health.
The NCFs under supply categories are as follows:
(43) Exploration and Extraction Of Fuels, (44) Fuel Refining and Processing Fuels,
(45) Generate Electricity, (46) Manufacture Equipment, (47) Produce and Provide Agricultural Products and Services, (48) Produce and Provide Human and Animal Food Products and Services, (49)Produce Chemicals, (50) Provide Metals and Materials, (51) Provide Housing, (52) Provide Information Technology Products and Services, (53) Provide Materiel and Operational Support to Defense,
(54) Research and Development, and (55) Supply Water.

\subsection{Semantic Scholar Open Research Corpus}
The Semantic Scholar Open Research Corpus (S2ORC) is a collection of open-access research papers and scholarly articles. This corpus was created by combining data from various sources and identifying open-access publications. It contains 136 million paper nodes, 76 million abstracts, and 12.7 million full-text papers. Medicine, biology, chemistry, engineering, computer science, physics, material science, math, psychology, economics, political science, business, geology, sociology, geography, environmental science, art, history, and philosophy are among the subjects covered by the corpus.
Each document in the S2ORC is available in JSON format, and each document has the following entries: paper ID, title, authors, abstract, year, arXiv ID, DOI, journal, field of study, and PDF parse (full PDF availability).

%% file: main.bbl
\begin{thebibliography}{10}
\providecommand{\url}[1]{#1}
\csname url@samestyle\endcsname
\providecommand{\newblock}{\relax}
\providecommand{\bibinfo}[2]{#2}
\providecommand{\BIBentrySTDinterwordspacing}{\spaceskip=0pt\relax}
\providecommand{\BIBentryALTinterwordstretchfactor}{4}
\providecommand{\BIBentryALTinterwordspacing}{\spaceskip=\fontdimen2\font plus
\BIBentryALTinterwordstretchfactor\fontdimen3\font minus
  \fontdimen4\font\relax}
\providecommand{\BIBforeignlanguage}[2]{{%
\expandafter\ifx\csname l@#1\endcsname\relax
\typeout{** WARNING: IEEEtran.bst: No hyphenation pattern has been}%
\typeout{** loaded for the language `#1'. Using the pattern for}%
\typeout{** the default language instead.}%
\else
\language=\csname l@#1\endcsname
\fi
#2}}
\providecommand{\BIBdecl}{\relax}
\BIBdecl

\bibitem{callaghan2020topography}
M.~W. Callaghan, J.~C. Minx, and P.~M. Forster, ``A topography of climate
  change research,'' \emph{Nature Climate Change}, vol.~10, no.~2, pp.
  118--123, 2020.

\bibitem{varini2020climatext}
F.~S. Varini, J.~Boyd-Graber, M.~Ciaramita, and M.~Leippold, ``{ClimaText: A}
  dataset for climate change topic detection,'' \emph{arXiv preprint
  arXiv:2012.00483}, 2020.

\bibitem{rolnick2022tackling}
D.~Rolnick, P.~L. Donti, L.~H. Kaack, K.~Kochanski, A.~Lacoste, K.~Sankaran,
  A.~S. Ross, N.~Milojevic-Dupont, N.~Jaques, A.~Waldman-Brown \emph{et~al.},
  ``Tackling climate change with machine learning,'' \emph{ACM Computing
  Surveys (CSUR)}, vol.~55, no.~2, pp. 1--96, 2022.

\bibitem{gao2020consistency}
M.~Gao, Z.~Zhang, G.~Yu, S.~{\"O}. Ar{\i}k, L.~S. Davis, and T.~Pfister,
  ``Consistency-based semi-supervised active learning: Towards minimizing
  labeling cost,'' in \emph{European Conference on Computer Vision}.\hskip 1em
  plus 0.5em minus 0.4em\relax Springer, 2020, pp. 510--526.

\bibitem{goudjil2018novel}
M.~Goudjil, M.~Koudil, M.~Bedda, and N.~Ghoggali, ``A novel active learning
  method using {SVM} for text classification,'' \emph{International Journal of
  Automation and Computing}, vol.~15, no.~3, pp. 290--298, 2018.

\bibitem{callaghan2021machine}
M.~Callaghan, C.-F. Schleussner, S.~Nath, Q.~Lejeune, T.~R. Knutson,
  M.~Reichstein, G.~Hansen, E.~Theokritoff, M.~Andrijevic, R.~J. Brecha
  \emph{et~al.}, ``Machine-learning-based evidence and attribution mapping of
  100,000 climate impact studies,'' \emph{Nature Climate Change}, vol.~11,
  no.~11, pp. 966--972, 2021.

\bibitem{ncfdefinition}
U.~D. of~Homeland~Security, ``{National Critical Functions: S}tatus update to
  the critical infrastructure community,'' 2020.

\bibitem{hsu2021diverse}
A.~Hsu and R.~Rauber, ``Diverse climate actors show limited coordination in a
  large-scale text analysis of strategy documents,'' \emph{Communications Earth
  \& Environment}, vol.~2, no.~1, pp. 1--12, 2021.

\bibitem{benites2018topic}
L.~Benites-Lazaro, L.~Giatti, and A.~Giarolla, ``Topic modeling method for
  analyzing social actor discourses on climate change, energy and food
  security,'' \emph{Energy Research \& Social Science}, vol.~45, pp. 318--330,
  2018.

\bibitem{chang2021applying}
I.-C. Chang, T.-K. Yu, Y.-J. Chang, and T.-Y. Yu, ``Applying text mining,
  clustering analysis, and latent dirichlet allocation techniques for topic
  classification of environmental education journals,'' \emph{Sustainability},
  vol.~13, no.~19, p. 10856, 2021.

\bibitem{marsi2014towards}
E.~Marsi, P.~{\O}zturk, E.~Aamot, G.~V. Sizov, and M.~V. Ardelan, ``Towards
  text mining in climate science: {E}xtraction of quantitative variables and
  their relations,'' 2014.

\bibitem{chen2019detecting}
X.~Chen, L.~Zou, and B.~Zhao, ``Detecting climate change deniers on twitter
  using a deep neural network,'' in \emph{Proceedings of the 2019 11th
  International Conference on Machine Learning and Computing}, 2019, pp.
  204--210.

\bibitem{ipcc2014}
I.~P. on~Climate~Change, ``{Climate Change 2014: Synthesis Report. Contribution
  of Working Groups I, II and III to the Fifth Assessment Report of the
  Intergovernmental Panel on Climate Change},'' 2014.

\bibitem{national2016attribution}
E.~National Academies~of Sciences, Medicine \emph{et~al.}, \emph{Attribution of
  extreme weather events in the context of climate change}.\hskip 1em plus
  0.5em minus 0.4em\relax National Academies Press, 2016.

\bibitem{task2021impacts}
T.~C. on~Future~Weather, C.~Extremes, M.~R. Tye, and J.~P. Giovannettone,
  ``Impacts of future weather and climate extremes on {United States}
  infrastructure: {A}ssessing and prioritizing adaptation actions,'' 2021.

\bibitem{usgcrp}
``{USGCRP Indicators Calalog},''
  \url{https://www.globalchange.gov/browse/indicators/catalog}, accessed:
  2022-10-19.

\bibitem{reidmiller2019fourth}
D.~Reidmiller, C.~Avery, D.~Easterling, K.~Kunkel, K.~Lewis, T.~Maycock, and
  B.~Stewart, ``Fourth national climate assessment,'' \emph{Volume II: Impacts,
  Risks, and Adaptation in the United States, Report-in-Brief}, 2019.

\bibitem{ipcc2022}
IPCC, ``{Climate Change 2022: Impacts, Adaptation and Vulnerability},'' 2022.

\bibitem{ncf}
``{National Critical Functions},''
  \url{https://www.cisa.gov/national-critical-functions-set}, accessed:
  2022-10-19.

\bibitem{lo-wang-2020-s2orc}
\BIBentryALTinterwordspacing
K.~Lo, L.~L. Wang, M.~Neumann, R.~Kinney, and D.~Weld, ``{S}2{ORC}: The
  semantic scholar open research corpus,'' in \emph{Proceedings of the 58th
  Annual Meeting of the Association for Computational Linguistics}.\hskip 1em
  plus 0.5em minus 0.4em\relax Online: Association for Computational
  Linguistics, 2020, pp. 4969--4983. [Online]. Available:
  \url{https://www.aclweb.org/anthology/2020.acl-main.447}
\BIBentrySTDinterwordspacing

\bibitem{reimers-2020-multilingual-sentence-bert}
\BIBentryALTinterwordspacing
N.~Reimers and I.~Gurevych, ``Making monolingual sentence embeddings
  multilingual using knowledge distillation,'' in \emph{Proceedings of the 2020
  Conference on Empirical Methods in Natural Language Processing}.\hskip 1em
  plus 0.5em minus 0.4em\relax Association for Computational Linguistics, 11
  2020. [Online]. Available: \url{https://arxiv.org/abs/2004.09813}
\BIBentrySTDinterwordspacing

\bibitem{devlin2018bert}
J.~Devlin, M.-W. Chang, K.~Lee, and K.~Toutanova, ``Bert: Pre-training of deep
  bidirectional transformers for language understanding,'' \emph{arXiv preprint
  arXiv:1810.04805}, 2018.

\bibitem{reimers2019sentence}
N.~Reimers and I.~Gurevych, ``{Sentence-BERT: Sentence embeddings using Siamese
  BERT-networks},'' \emph{arXiv preprint arXiv:1908.10084}, 2019.

\bibitem{ruhling2021end}
S.~R{\"u}hling~Cachay, B.~Boecking, and A.~Dubrawski, ``End-to-end weak
  supervision,'' \emph{Advances in Neural Information Processing Systems},
  vol.~34, 2021.

\bibitem{bach2017weaksupervision}
S.~H. Bach, B.~He, A.~Ratner, and C.~R{\'e}, ``Learning the structure of
  generative models without labeled data,'' in \emph{International Conference
  on Machine Learning}.\hskip 1em plus 0.5em minus 0.4em\relax PMLR, 2017, pp.
  273--282.

\bibitem{ratner2017snorkel}
A.~Ratner, S.~H. Bach, H.~Ehrenberg, J.~Fries, S.~Wu, and C.~R{\'e}, ``Snorkel:
  Rapid training data creation with weak supervision,'' in \emph{Proceedings of
  the VLDB Endowment. International Conference on Very Large Data Bases},
  vol.~11, no.~3.\hskip 1em plus 0.5em minus 0.4em\relax NIH Public Access,
  2017, p. 269.

\bibitem{reimers-2019-sentence-bert}
\BIBentryALTinterwordspacing
N.~Reimers and I.~Gurevych, ``{Sentence-BERT: Sentence Embeddings using Siamese
  BERT-Networks},'' in \emph{Proceedings of the 2019 Conference on Empirical
  Methods in Natural Language Processing}.\hskip 1em plus 0.5em minus
  0.4em\relax Association for Computational Linguistics, 11 2019. [Online].
  Available: \url{https://arxiv.org/abs/1908.10084}
\BIBentrySTDinterwordspacing

\bibitem{liu2019roberta}
Y.~Liu, M.~Ott, N.~Goyal, J.~Du, M.~Joshi, D.~Chen, O.~Levy, M.~Lewis,
  L.~Zettlemoyer, and V.~Stoyanov, ``Roberta: {A} robustly optimized bert
  pretraining approach,'' \emph{arXiv preprint arXiv:1907.11692}, 2019.

\bibitem{read2011classifier}
J.~Read, B.~Pfahringer, G.~Holmes, and E.~Frank, ``Classifier chains for
  multi-label classification,'' \emph{Machine Learning}, vol.~85, no.~3, pp.
  333--359, 2011.

\bibitem{blei2003latent}
D.~M. Blei, A.~Y. Ng, and M.~I. Jordan, ``Latent {D}irichlet allocation,''
  \emph{Journal of Machine Learning Research}, vol.~3, no. Jan., pp. 993--1022,
  2003.

\bibitem{grootendorst2022bertopic}
M.~Grootendorst, ``{BERTopic: Neural topic modeling with a class-based TF-IDF
  procedure},'' \emph{arXiv preprint arXiv:2203.05794}, 2022.

\bibitem{mcinnes2018umap}
L.~McInnes, J.~Healy, and J.~Melville, ``Umap: Uniform manifold approximation
  and projection for dimension reduction,'' \emph{arXiv preprint
  arXiv:1802.03426}, 2018.

\bibitem{mcinnes2017hdbscan}
L.~McInnes, J.~Healy, and S.~Astels, ``hdbscan: Hierarchical density based
  clustering.'' \emph{J. Open Source Softw.}, vol.~2, no.~11, p. 205, 2017.

\bibitem{Ratner2022}
A.~Ratner and H.~Ehrenberg, ``Snorkel,''
  \url{https://github.com/snorkel-team/snorkel}, 2022.

\end{thebibliography}
